\documentclass[conference]{IEEEtran}
\IEEEoverridecommandlockouts
\usepackage{cite}
\usepackage{amsmath,amssymb,amsfonts}
\usepackage{algorithmic}
\usepackage{graphicx}
\usepackage{textcomp}
\usepackage{xcolor}
\usepackage{booktabs}
\usepackage{tabularx}
\usepackage[ruled,vlined,linesnumbered]{algorithm2e}

\def\BibTeX{{\rm B\kern-.05em{\sc i\kern-.025em b}\kern-.08em T\kern-.1667em\lower.7ex\hbox{E}\kern-.125emX}}

\begin{document}
	
\title{TimeGNN-Augmented Hybrid-Action MARL for Fine-Grained Task Partitioning and Energy-Aware Offloading in MEC}

\author{
	\IEEEauthorblockN{Wei Ai}
	\IEEEauthorblockA{\textit{College of Computer and Mathematics} \\
		\textit{Central South University of Forestry and Technology}\\
		ChangSha, China \\
		aiwei@hnu.edu.cn}
	\and
	\IEEEauthorblockN{Yun Peng}
	\IEEEauthorblockA{\textit{College of Computer and Mathematics} \\
		\textit{Central South University of Forestry and Technology}\\
		ChangSha, China \\
		pengyun@csuft.edu.cn}
	\and
	\IEEEauthorblockN{Yuntao Shou}
	\IEEEauthorblockA{\textit{College of Computer and Mathematics} \\
		\textit{Central South University of Forestry and Technology}\\
		ChangSha, China \\
		yuntao\_shou@csuft.edu.cn}
	\and
	\IEEEauthorblockN{\hspace{5em}Tao Meng{*}}
		\IEEEauthorblockA{\textit{College of Computer and Mathematics}\\
		\textit{\hspace{4em}Central South University of Forestry and Technology}\\
		\hspace{4em}ChangSha, China \\
		\hspace{4em}mengtao@hnu.edu.cn}
	\and
	\IEEEauthorblockN{\hspace{2em}Keqin Li}
	\IEEEauthorblockA{\hspace{3em}Department of Computer Science \\
		\hspace{3em}State University of New York\\
		\hspace{3em}New Paltz, New York 12561, USA \\
		\hspace{4em}lik@newpaltz.edu}
	\thanks{* is the corresponding author.}
}
	
\maketitle
	
\begin{abstract}
With the rapid growth of IoT devices and latency-sensitive applications, the demand for both real-time and energy-efficient computing has surged, placing significant pressure on traditional cloud computing architectures. Mobile edge computing (MEC), an emerging paradigm, effectively alleviates the load on cloud centers and improves service quality by offloading computing tasks to edge servers closer to end users. However, the limited computing resources, non-continuous power provisioning (e.g., battery-powered nodes), and highly dynamic systems of edge servers complicate efficient task scheduling and resource allocation. To address these challenges, this paper proposes a multi-agent deep reinforcement learning algorithm, TG-DCMADDPG, and constructs a collaborative computing framework for multiple edge servers, aiming to achieve joint optimization of fine-grained task partitioning and offloading. This approach incorporates a temporal graph neural network (TimeGNN) to model and predict time series of multi-dimensional server state information, thereby reducing the frequency of online interactions and improving policy predictability. Furthermore, a multi-agent deterministic policy gradient algorithm (DC-MADDPG) in a discrete-continuous hybrid action space is introduced to collaboratively optimize task partitioning ratios, transmission power, and priority scheduling strategies. Extensive simulation experiments confirm that TG-DCMADDPG achieves markedly faster policy convergence, superior energy–latency optimization, and higher task completion rates compared with existing state-of-the-art methods, underscoring its robust scalability and practical effectiveness in dynamic and constrained MEC scenarios.
\end{abstract}
	

\begin{figure}[t]         
    \centering
    \includegraphics[width=\linewidth]{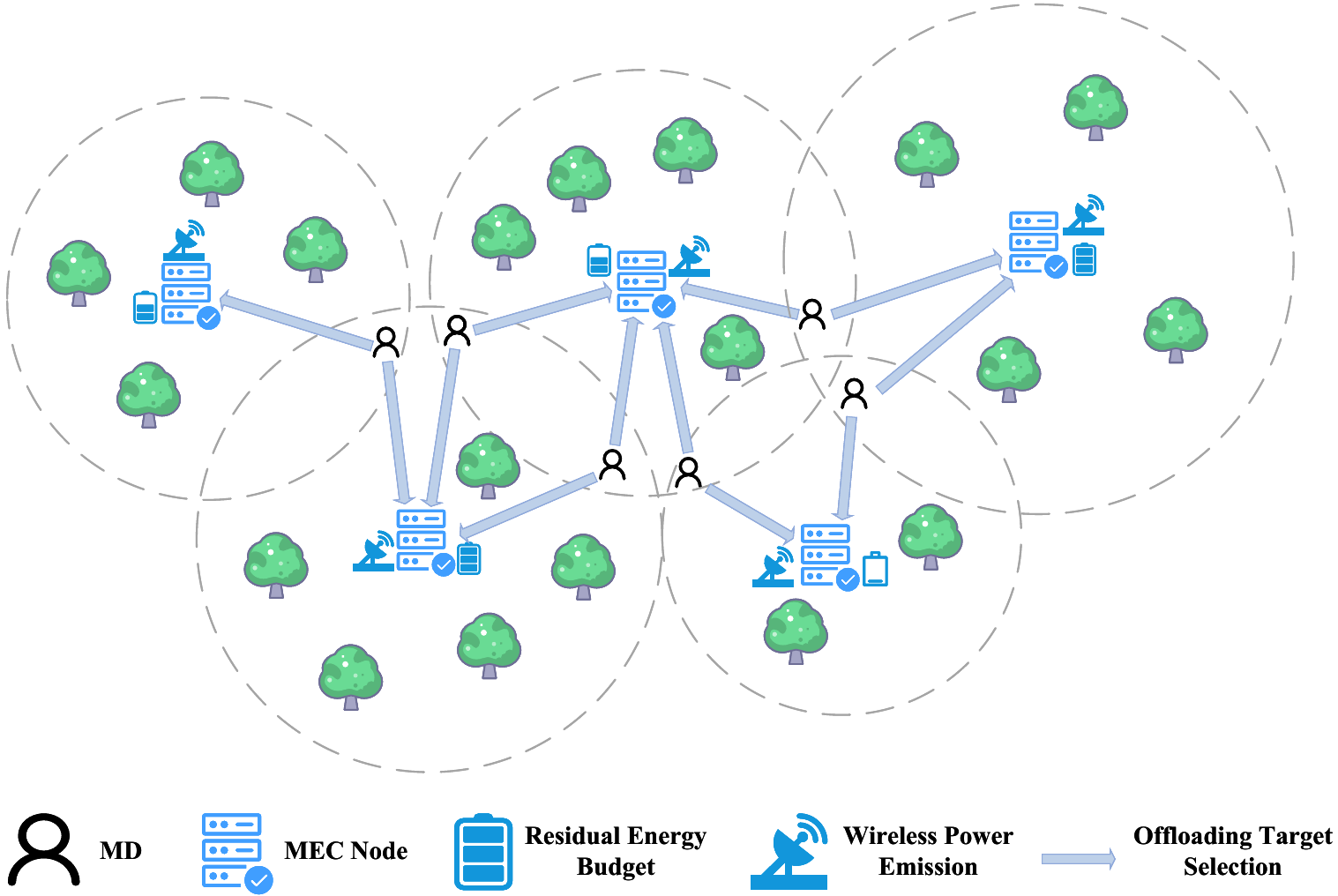}
    \caption{Illustrates the considered MEC system, where each edge server operates on battery power without a stable power supply. To prolong service availability and achieve load balancing, a time-series prediction model is employed to forecast the state of edge devices. Based on these predictions, task partitioning and offloading strategies are applied to reduce energy consumption and enhance QoS.}
    \label{fig:Environment}
\end{figure}

\section{Introduction}
The proliferation of IoT devices and latency-sensitive applications~\cite{industrial-mec, comprehensive2025, Conversational2022, Adversarial2024, low-rank2024} has triggered immense computation demand, which traditional cloud services struggle to meet due to high latency and energy costs~\cite{vu2018offloading, Object2022, Der-gcn2024, Masked2025, Deep2024}. Mobile Edge Computing (MEC) has emerged as an effective solution by relocating compute resources closer to end users~\cite{yang2023energy, multi-message2024, Spegcl2025, alignment2024}. 

Existing MEC scheduling approaches fall into two broad categories: optimization-based methods, such as mixed-integer nonlinear programming (MINLP) for joint latency–energy minimization~\cite{vu2018offloading, Efficient2024, Contrastive2025, bottleneck2025}, and learning-driven paradigms. Particularly, multi-agent reinforcement learning (MARL) frameworks like MADDPG, based on the CTDE architecture, have enabled decentralized offloading decisions~\cite{jiang2024task,ctde2025overview, Graphunet2023}. Many of these approaches incorporate energy-awareness through policies that adapt to energy-harvesting capabilities~\cite{hierarchicalMARL2023, adaptation2025, Dynamic2025}.

However, MEC nodes in remote or intermittently powered environments face stringent battery constraints, limiting operational sustainability and making frequent agent-server exchanges impractical. To mitigate this, proactive mechanisms that forecast future server states could reduce interaction overhead and sustain performance. In this vein, temporal graph neural networks (TimeGNN) have shown effective modeling of multi-variate time-series dynamics with scalable inference~\cite{xu2023TimeGNN, diffusion2026, survey2025}. More generally, spatiotemporal GNNs are gaining traction in forecasting tasks across domains~\cite{surveyGNN4TS, GSDNet2025, CILF-CIAE2025}. Moreover, workload dynamics and user mobility introduce load imbalance, where static offloading policies fail to ensure efficient resource use; distributed task slicing that accounts for these dynamics is crucial~\cite{MEHR2021, Flow2025}.  

Motivated by these challenges, this paper presents TG-DCMADDPG, a novel hybrid framework combining TimeGNN-based state prediction with a discrete–continuous action MARL approach, enabling predictive, energy-aware, and fine-grained offloading suitable for heterogeneous and constrained MEC environments.

\begin{itemize}
    \item  A predictive cooperative MEC offloading framework is proposed, combining time-series prediction and multi-agent learning to address task splitting, resource allocation, and energy efficiency under dynamic server conditions.resource allocation, and energy efficiency under dynamic server conditions.
    
    \item  A TimeGNN-based state predictor is designed to capture the temporal evolution of multi-dimensional edge server states, enabling proactive and adaptive offloading decisions while reducing online interaction overhead.
    
    \item  A hybrid-action DC-MADDPG algorithm is developed to jointly optimize continuous and discrete actions under a centralized training–decentralized execution paradigm, achieving load-balanced, low-latency, energy-efficient task offloading across multiple edge servers.
\end{itemize}

\section{Related Work}

\subsection{Energy-Constrained Task Offloading in MEC}

In mobile edge computing (MEC), task offloading is a fundamental technique to reduce computational delay and energy consumption, especially when edge servers operate with limited battery capacity or rely on intermittent energy sources~\cite{b1,yang2023energy,vu2018offloading,MEHR2021}.  
Traditional approaches are predominantly optimization-based, where the offloading and resource allocation problem is formulated as a mixed-integer nonlinear program (MINLP) or convex optimization model~\cite{b2,b3,b4}.  
For example, Mao \emph{et al.}~\cite{b1} modeled the process as a Markov Decision Process (MDP) to jointly minimize delay and energy cost, while Chen \emph{et al.}~\cite{b2} proposed a convex optimization framework for multi-user MEC under bandwidth constraints.  
Extensions to energy-harvesting scenarios include dynamic voltage scaling for partial offloading~\cite{b3} and cooperative energy-aware offloading leveraging user mobility~\cite{vu2018offloading,MEHR2021}.  
Although these methods can achieve optimal solutions in static environments, their dependence on accurate environment models limits adaptability to dynamic network conditions, heterogeneous resources, and long-term energy constraints.

\subsection{DRL-based Predictive Resource Management with Hybrid Actions}

Deep reinforcement learning (DRL) has gained traction as a model-free alternative capable of learning adaptive offloading strategies in dynamic MEC environments~\cite{b5,b6,jiang2024task,industrial-mec}.  
Single-agent DRL approaches such as DROO~\cite{b5} and LSTM-DQN~\cite{b6} improve adaptability but face scalability challenges in multi-device systems.  
Multi-agent reinforcement learning (MARL) frameworks, including Com-DDPG~\cite{b7} and energy-aware MARL strategies~\cite{b8,d2d-marl}, enable decentralized decision-making by modeling interactions among multiple mobile devices.  
Recent advances explore hybrid action spaces that combine discrete decisions (e.g., server selection) with continuous controls (e.g., task splitting ratio, transmission power)~\cite{sagin,hybrid-marl}, enabling fine-grained resource allocation.  
Moreover, temporal graph neural networks (TimeGNN)~\cite{xu2023TimeGNN} have demonstrated strong capability in modeling spatiotemporal dependencies for multivariate time-series forecasting.  
Integrating predictive models like TimeGNN into MARL can reduce costly agent–server interactions and improve decision-making foresight.  
However, the integration of predictive temporal modeling with hybrid-action MARL in energy-constrained MEC remains limited, and existing works often remain reactive, resulting in suboptimal load balancing and accelerated energy depletion under volatile network conditions.

In general, existing research on MEC task scheduling has achieved substantial progress in both energy-constrained optimization and learning-based resource management. Optimization frameworks offer provably optimal or near-optimal performance under static or well-characterized conditions, while learning-based approaches,particularly MARL,excel in handling dynamic and partially observable environments. Nevertheless, most studies treat system state acquisition as an instantaneous process, overlooking the overhead of frequent environment interactions and the potential benefits of predictive modeling. Moreover, the integration of temporal graph neural networks with hybrid-action MARL for proactive, fine-grained task partitioning under long-term energy constraints remains largely unexplored. These gaps motivate our TG-DCMADDPG framework, which bridges predictive spatiotemporal modeling and distributed hybrid-action decision-making to enable energy-aware, forward-looking MEC task offloading.

\section{System Model}

    \subsection{System Overview}

    As illustrated in Fig.~\ref{fig:Environment}, the proposed mobile edge computing (MEC) framework adopts a hierarchical architecture comprising a mobile device (MD) layer and an edge server (MEC) layer. The system operates in discrete time slots \( T = \{0, 1, \dots, t-1\} \), each of duration \( \Delta \), during which task offloading and resource allocation are dynamically optimized.
    
    A set of mobile devices \( M = \{1, 2, \dots, M\} \) move within the service area and periodically generate latency-sensitive, compute-intensive tasks. To mitigate local resource constraints and reduce processing delay, each device can offload its task to one or two available MEC servers \( N = \{1, 2, \dots, N\} \) within communication range. Each MEC node has limited computing capacity, bandwidth, and energy resources, necessitating energy-aware and delay-sensitive coordination.
    
    The goal of the system is to jointly optimize task offloading, transmission power, and computation resource allocation to minimize overall energy consumption and execution delay under dynamic network conditions.

	\subsection{Communication Model}
	We use a general task model $T_n = \{\lambda_n, \text{PRIn}\}$, where $\lambda_n$ represents the input data size of task $T_n$, and $\text{PRIn}$ represents the priority of the task. The higher the priority, the higher the CPU frequency allocated from the remaining resources of the edge server. The allocated frequency $C_n$ can be expressed as:
	\begin{equation}
		\rho = \beta \times \text{PRIn}
	\end{equation}
	where \(\beta\) is a tunable scaling factor that reflects the optimization granularity of frequency assignment.

	Based on the general communication model, The data transmission rate between MEC node $m$ and mobile user $n$ follows the Shannon capacity model and is given by:
    \begin{equation}
        r(m, n) = W_m \log_2 \!\left( 1 + \frac{P_n H_{m,n}}{N_0 W_m} \right)
    \end{equation}
    where $W_m$ denotes the channel bandwidth allocated by MEC node $m$ to mobile user $n$, 
    $P_n$ is the transmission power of mobile user $n$, 
    $H_{m,n}$ represents the channel gain, 
    and $N_0$ is the noise power spectral density. 
    Therefore, the input data transmission time can be expressed as:
	\begin{equation}
		t_{\text{trans}}(m, n) = \frac{\alpha \lambda_n}{r(m, n)}
	\end{equation}
	where $\alpha$ represents the proportion of the task that mobile user $n$ offloads to one of the edge servers, and $m1$ and $m2$ represent the two MEC nodes to which user $n$ offloads the task. $r(m1, n)$ and $r(m2, n)$ represent the transmission rates between user $n$ and edge servers $m1$ and $m2$, respectively. The total energy consumed during transmission is:
	\begin{equation}
		E_{\text{trans}}(n) = P_n t_{\text{trans}}(m, n)
	\end{equation}
	
	\subsection{Computation Model}
	As discussed previously, each mobile device offloads a portion of its task to nearby MEC servers for remote execution. We assume that MEC nodes are capable of processing multiple tasks in parallel through processor sharing. Let the maximum computational capacity of edge server \( m \) be \( f_m \), and let its current workload be \( L_m \), representing the total amount of CPU frequency already allocated to other tasks during the current time slot. The normalized available computing capacity is thus \( 1 - \frac{L_m}{f_m} \).
	
	The actual CPU frequency allocated by server \( m \in \{m_1, m_2\} \) to user \( n \) is given by:
	\begin{equation}
		f(m, n) = \rho_{m,n} \cdot f_m \left(1 - \frac{L_m}{f_m}\right)
	\end{equation}
	where \( \rho_{m,n} \in [0,1] \) denotes the proportion of the available computing capacity assigned to task \( n \). 
	
	The power consumption for computing the task is then expressed as:
	\begin{equation}
		P_{\text{comp}}(m, n) = \kappa \cdot f(m, n)^3
	\end{equation}
	where \( \kappa \) is a hardware-dependent effective switched capacitance coefficient, 
	
	Let \( \alpha_{m_1} \) and \( \alpha_{m_2} \) represent the proportions of the task offloaded by user \( n \) to MEC servers \( m_1 \) and \( m_2 \), respectively, satisfying \( \alpha_{m_1} + \alpha_{m_2} = 1 \). The execution delay for the partial task assigned to server \( m \) is:
	\begin{equation}
		t_{\text{comp}}(m, n) = \frac{\alpha_m \cdot \lambda_n}{f(m, n)}
	\end{equation}
	
	The corresponding computational energy consumption is:
	\begin{equation}
		E_{\text{comp}}(m, n) = P_{\text{comp}}(m, n) \cdot t_{\text{comp}}(m, n)
	\end{equation}
	
	the total energy consumption incurred by task \( n \) across both MEC nodes is:
	\begin{equation}
    \begin{aligned}
		E_{\text{total}}(n) &= E_{\text{trans}}(m_1, n) + E_{\text{trans}}(m_2, n) \\&+ E_{\text{comp}}(m_1, n) + E_{\text{comp}}(m_2, n)
        \end{aligned}
	\end{equation}
	
	Since the task is split and executed in parallel, the overall execution delay is determined by the slower processing branch:
	\begin{equation}
		t_{\text{total}}(n) = \max\left(t_{\text{comp}}(m_1, n), t_{\text{comp}}(m_2, n)\right)
	\end{equation}
	
	This model captures the impact of task splitting, MEC server workload, and task priority on computation delay and energy consumption in a distributed edge computing environment.

    \section{Problem Formulation}
    
    Based on the system model described above, we formulate the joint optimization problem of task offloading and resource allocation in a multi-access edge computing (MEC) system. The objective is to minimize the overall system cost by dynamically optimizing the offloading ratio, transmission power, and computing resource allocation for all mobile devices (MDs).
    
    At each time slot \(t \in \{0, 1, \dots, T-1\}\), each mobile device \(n \in \mathcal{N}\) generates a latency-sensitive task. According to the system model, the task may be partitioned and offloaded to one or two reachable MEC servers for execution. Let \(\alpha_{m,n}(t)\), \(P_n(t)\), and \(f(m,n,t)\) denote the offloading ratio, transmission power, and allocated CPU frequency, respectively.
    
    The instantaneous system cost for device \(n\) is defined as a weighted sum of energy consumption and execution delay.
    \begin{equation}
        \mathcal{C}_n(t) = \omega_E E_{\text{total}}(n,t) + \omega_D t_{\text{total}}(n,t)
    \end{equation}
    where \(\omega_E\) and \(\omega_D\) represent the relative importance of energy efficiency and latency reduction.
    
    The long-term optimization problem is formulated as:
    \begin{equation}
        \min_{\{\alpha, P, f\}} \quad 
        \mathbb{E}\!\left[\sum_{t=0}^{T-1} \sum_{n \in \mathcal{N}} \mathcal{C}_n(t)\right]
    \end{equation}
    subject to the following constraints:

	\subsubsection*{1) Task Offloading Constraints}
	Each task must be fully offloaded, with valid ratio bounds:
	\begin{equation}
		\sum_{m \in \mathcal{M}} \alpha_{m,n} = 1, \quad 0 \leq \alpha_{m,n} \leq 1,\quad \forall n \in \mathcal{N}
	\end{equation}
	
	\subsubsection*{2) Computation Resource Constraints}
	Each MEC server has limited computing capacity at time \(t\):
	\begin{equation}
		\sum_{n \in \mathcal{N}} f(m,n,t) \leq f_m,\quad \forall m \in \mathcal{M}
	\end{equation}
	
	\subsubsection*{3) Transmission Power Constraints}
	The transmission power must support offloading given channel conditions:

	\begin{equation}
		P_n \geq \frac{N_0 D_{m}^2}{W_m H_{m,n}}, \quad \forall n \in \mathcal{N},\ \forall m \in \mathcal{M}
	\end{equation}
	
	\subsubsection*{4) Server Energy Budget Constraints}
	Each MEC node \(m\) must operate within its finite energy budget \(E_m^{\max}\):
	\begin{equation}
		\sum_{t=0}^{T-1} \sum_{n \in \mathcal{N}} E_{\text{comp}}(m,n,t) \leq E_m^{\max},\quad \forall m \in \mathcal{M}
	\end{equation}
	
	\subsubsection*{5) Resource Coupling Constraint}
	The computing frequency assigned by server \(m\) should match the task portion it handles:
	\begin{equation}
		f(m,n,t) \propto \alpha_{m,n}(t),\quad \forall m \in \mathcal{M},\ \forall n \in \mathcal{N}
	\end{equation}

	\subsubsection*{6) Offloading Range Constraint}
	Only reachable MEC servers are considered for task offloading. Specifically, the physical distance between user \(n\) and server \(m\) must not exceed the maximum transmission range \(V_\pi\), which is determined by transmission power and environmental conditions:
	\begin{equation}
		d_{m,n} \leq V_{\pi},\quad \forall m \in \mathcal{M},\ \forall n \in \mathcal{N}
	\end{equation}
	
	The maximum communication range \(V_\pi\) is further constrained by the minimum required received power \(P_{\text{th}}\)
	\begin{equation}
		V_\pi = \left( \frac{P_n G_t G_r \lambda^2}{(4\pi)^2 P_{\text{th}}} \right)^{\frac{1}{\eta}}
	\end{equation}
	where \(P_n\) is the transmission power of user \(n\), \(G_t\) and \(G_r\) are antenna gains, \(\lambda\) is the carrier wavelength, and \(\eta\) is the path-loss exponent.
	\subsection*{Problem Characteristics}
	
	The formulated optimization problem is a constrained mixed-integer nonlinear programming (MINLP) problem involving continuous control variables (\(\alpha, P, f\)) and dynamic temporal coupling across time slots. The stochastic nature of task generation, channel fading, and server loads leads to a high-dimensional and partially observable state space.
	
	To tackle these challenges, we adopt a distributed multi-agent reinforcement learning (MARL) framework. Specifically, we design a temporal graph-based state encoder (TimeGNN) to model spatiotemporal dependencies and integrate it with a centralized-critic actor–critic algorithm (DC-MADDPG), enabling scalable and cooperative decision-making among MDs
	
	\section{Proposed Algorithm Framework: TG-DCMADDPG}
	\begin{figure*}[t]         
		\centering
		\includegraphics[width=\textwidth]{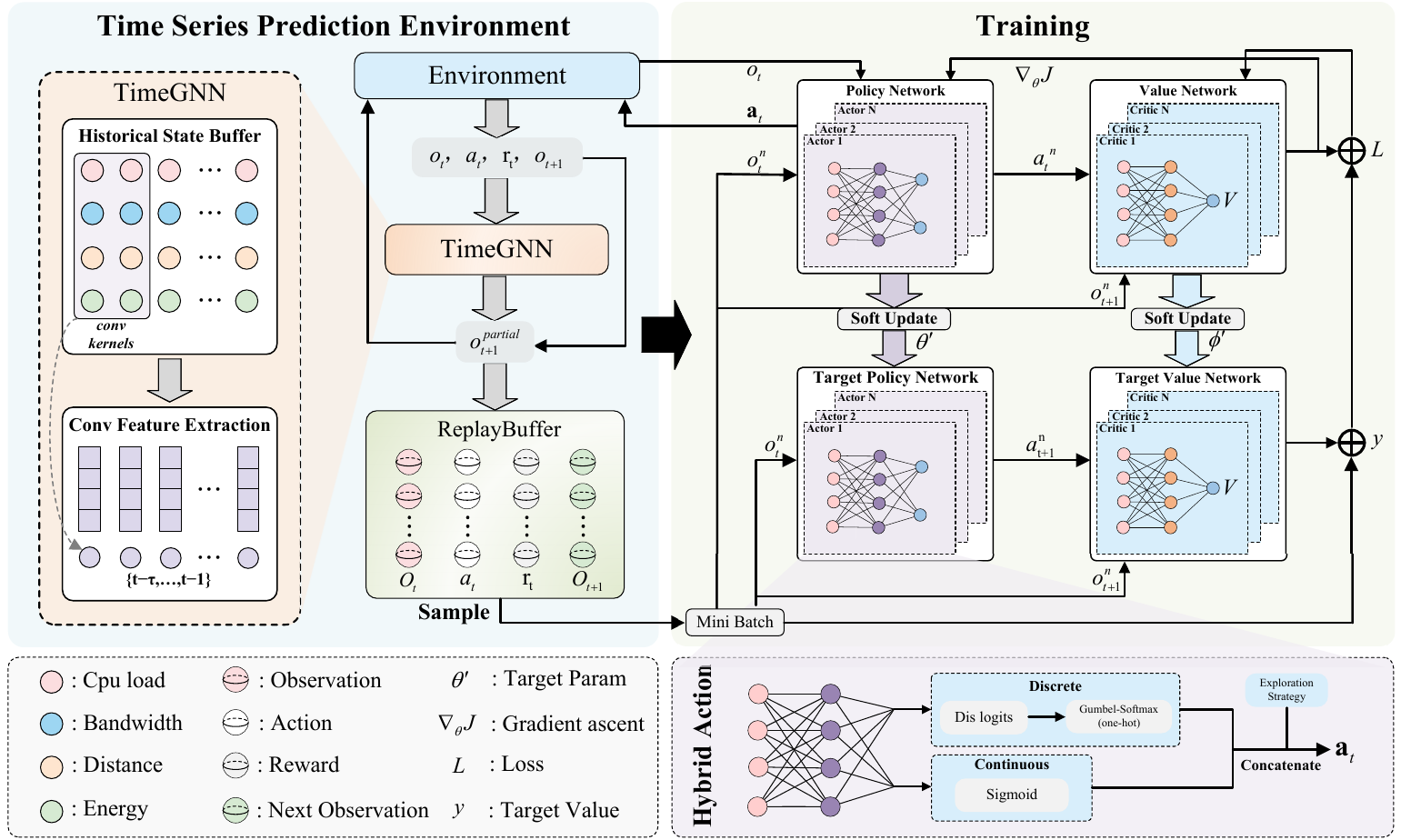}
		\caption{presents the TG-DCMADDPG framework, which couples TimeGNN-based state prediction with hybrid-action MARL for energy-aware MEC task partitioning. Historical states are transformed into temporal graphs, enabling future server state forecasting. Augmented observations feed a CTDE training loop, where actors output discrete–continuous actions for server selection, task splitting, and power control, while critics guide updates via soft target networks. This integration enables proactive, fine-grained offloading with reduced interaction overhead and improved latency–energy trade-offs in dynamic edge environments.}
		\label{fig:algorithm}
        \vspace{-2mm}
	\end{figure*}
	
	\subsection{Overview}
	
	To address the dual challenges of energy sustainability and intelligent task scheduling in edge computing environments with intermittent power supply, we propose a novel learning-based decision framework, termed {TG-DCMADDPG}. The framework integrates temporal graph neural networks (TimeGNN) for state prediction and a discrete-continuous multi-agent DDPG algorithm (DC-MADDPG) for cooperative task offloading and resource allocation. This design jointly addresses temporal uncertainty, hybrid decision spaces, and long-term energy-aware optimization in dynamic MEC systems.

    \subsection{Predictive State Augmentation via TimeGNN}
    
    Accurately capturing the future evolution of MEC server states (e.g., CPU load, remaining energy, and bandwidth) is critical for reducing agent–server interaction overhead and enabling proactive decision-making. To this end, We use a temporal graph neural network (TimeGNN) to jointly model the spatial correlations between servers and the temporal dependencies in their state trajectories.
    
    At each time slot $t$, the MEC network is represented as a dynamic graph $\mathcal{G}_t = (\mathcal{V}, \mathcal{E}_t)$, where each node $v \in \mathcal{V}$ corresponds to an edge server, and edge weights in $\mathcal{E}_t$ encode time-varying network connectivity or communication quality. Let $H_t^{(l)} \in \mathbb{R}^{|\mathcal{V}|\times d_l}$ denote the $l$-th layer node embeddings. TimeGNN updates node representations via a spatial–temporal message passing mechanism:
    \begin{equation}
        H_t^{(l+1)} = \sigma \left( \tilde{A}_t H_t^{(l)} W_s^{(l)} + \Phi_t \odot \left(H_t^{(l)} W_t^{(l)}\right) \right),
        \label{eq:timegnn_update}
    \end{equation}
    where $\tilde{A}_t$ is the normalized adjacency matrix capturing spatial relationships, $\Phi_t$ is a temporal encoding derived from a recurrent unit (e.g., GRU) that aggregates multi-step historical information, $W_s^{(l)}$ and $W_t^{(l)}$ are learnable spatial and temporal projection matrices, and $\sigma(\cdot)$ is a nonlinear activation (ReLU in our implementation).
    
    After $L$ layers, the final node embeddings $H_t^{(L)}$ are passed through a multi-layer perceptron (MLP) to generate the $\Delta$-step-ahead prediction:
    \begin{equation}
        \hat{s}_{t+\Delta} = \mathrm{MLP}(H_t^{(L)}),
        \label{eq:timegnn_predict}
    \end{equation}
    where $\hat{s}_{t+\Delta}$ encodes the anticipated CPU utilization, residual energy, and link quality for each server at time $t+\Delta$. These predicted states are concatenated with the current observation of agent $n$:
    \begin{equation}
        \tilde{o}_t^n = \left[o_t^n, \hat{s}_{t+\Delta}\right],
        \label{eq:aug_obs}
    \end{equation}
    producing an augmented observation that enables agents to make forward-looking decisions without frequent real-time state queries, thereby lowering communication and computation overhead.
    
    \subsection{Hybrid Action Learning via DC-MADDPG}
    
    Given the augmented observation $\tilde{o}_t^n$, we adopt a discrete–continuous multi-agent deep deterministic policy gradient (DC-MADDPG) framework to learn hybrid offloading strategies under the centralized training and decentralized execution (CTDE) paradigm. Each agent’s policy network outputs:
    \begin{equation}
        a_t^n = \left[ a_t^{n,d}, a_t^{n,c} \right] \in \mathbb{R}^5,
        \label{eq:hybrid_action}
    \end{equation}
    where $a_t^{n,d} \in \mathbb{R}^3$ is a differentiable one-hot vector representing server selection (sampled via Gumbel–Softmax for backpropagation), and $a_t^{n,c} \in \mathbb{R}^2$ specifies the task-splitting ratio and transmission power level.

    During training, the centralized critic $Q_{\phi}(\mathbf{s}_t, \mathbf{a}_t)$ evaluates the joint action $\mathbf{a}_t = \{a_t^n\}_{n=1}^N$ given the global state $\mathbf{s}_t$, while each actor $\mu_\theta^n$ updates its policy using only local augmented observations. The TD target for critic optimization is:
    \begin{equation}
        y_t = r_t + \gamma Q_{\phi'}(\mathbf{s}_{t+1}, \mu_{\theta'}(\mathbf{s}_{t+1})),
        \label{eq:td_target}
    \end{equation}
    where $\phi'$ and $\theta'$ are target network parameters. The critic is trained by minimizing:
    \begin{equation}
        \mathcal{L}_Q = \mathbb{E} \left[ \left( Q_{\phi}(\mathbf{s}_t, \mathbf{a}_t) - y_t \right)^2 \right],
        \label{eq:td_loss}
    \end{equation}
    and the actor is updated via the deterministic policy gradient:
    \begin{equation}
        \nabla_{\theta} J = \mathbb{E} \left[ \nabla_a Q_{\phi}(\mathbf{s}_t, \mathbf{a}_t) \big|_{\mathbf{a}_t=\mu_{\theta}(\mathbf{s}_t)} \cdot \nabla_{\theta} \mu_{\theta}(\mathbf{s}_t) \right].
        \label{eq:policy_grad}
    \end{equation}
    To ensure stability, target networks are updated softly:
    \begin{equation}
        \theta' \leftarrow \tau \theta + (1 - \tau)\theta', \quad \phi' \leftarrow \tau \phi + (1 - \tau)\phi'.
        \label{eq:soft_update}
    \end{equation}
    
    By integrating predictive state modeling with hybrid-action MARL, the proposed TG-DCMADDPG enables foresighted coordination of task offloading, transmission power control, and workload partitioning, thereby improving energy efficiency and load balancing under dynamic and resource-constrained MEC environments. 
    Algorithm~\ref{alg:tg-dcmaddpg} outlines the overall training process of TG-DCMADDPG.

    
    
    \begin{algorithm}[htbp]
\caption{TG-DCMADDPG with Dual Buffers for Energy-Aware MEC Task Offloading}
\label{alg:tg-dcmaddpg}
\KwIn{
    RL replay buffer $\mathcal{D}_{\mathrm{RL}}$, 
    temporal buffer $\mathcal{B}_{\mathrm{Temp}}$ (length $K_{\text{hist}}$), 
    target update rate $\tau$, 
    prediction horizon $\Delta$, 
    prediction period $K_{\text{pred}}$
}
\KwOut{Trained actor $\mu_{\theta}$ and critic $Q_{\phi}$}

\For{episode $= 1$ to $E_{\max}$}{
    Initialize environment; clear $\mathcal{B}_{\mathrm{Temp}}$\;
    Observe initial $o_0^n$ for each agent $n$ and append to $\mathcal{B}_{\mathrm{Temp}}$\;
    
    \For{$t = 0$ to $T-1$}{
        \tcp{--- TimeGNN Prediction using Temporal Buffer ---}
        \If{$t \bmod K_{\text{pred}} = 0$}{
            Construct server graph $\mathcal{G}_t$ from $\mathcal{B}_{\mathrm{Temp}}$\;
            Predict $\hat{s}_{t+\Delta}$ via Eqs.~\eqref{eq:timegnn_update}--\eqref{eq:timegnn_predict}\;
        }
        Augment observation $\tilde{o}_t^n \leftarrow [o_t^n, \hat{s}_{t+\Delta}]$ (Eq.~\eqref{eq:aug_obs})\;
        
        \tcp{--- Hybrid Action Selection via Actor ---}
        Each agent selects $a_t^n$ using Eq.~\eqref{eq:hybrid_action} from $\mu_{\theta}$\;
        Execute $\mathbf{a}_t$, observe $r_t$ and next $o_{t+1}^n$\;
        
        \tcp{--- Buffer Updates ---}
        Append $o_{t+1}^n$ to $\mathcal{B}_{\mathrm{Temp}}$ (FIFO to keep last $K_{\text{hist}}$ steps)\;
        Store $(\mathbf{o}_t, \mathbf{a}_t, r_t, \mathbf{o}_{t+1})$ into $\mathcal{D}_{\mathrm{RL}}$\;
        
        \tcp{--- Centralized Training of DC-MADDPG ---}
        Sample minibatch from $\mathcal{D}_{\mathrm{RL}}$\;
        Update critic by minimizing Eq.~\eqref{eq:td_loss} with target Eq.~\eqref{eq:td_target}\;
        Update actor via Eq.~\eqref{eq:policy_grad}\;
        Soft-update target networks via Eq.~\eqref{eq:soft_update}\;
    }
}
\end{algorithm}
\vspace{-2mm}

	\section{Simulation Results and Discussion}

    \subsection{Experimental Setup}

    We evaluate the performance of the proposed TG-DCMADDPG algorithm through extensive simulations and comparisons with baseline methods. All experiments are implemented in Python 3.8.10 using the PyTorch 1.8.1 framework and executed on an NVIDIA RTX 4060 GPU.
    
    A dynamic mobile edge computing (MEC) environment is simulated where mobile devices are randomly distributed and generate latency-sensitive tasks over time. At each time slot, each user may offload its task to any two edge servers selected from a set of 3, 5, 10, or 15 servers, depending on the scenario. Wireless channel conditions and user locations vary temporally, introducing stochastic dynamics into the system. 
    
    For the TimeGNN module, the model follows the architecture in~\cite{xu2023TimeGNN}, consisting of three stacked GraphSAGE layers for temporal graph reasoning. It takes as input the past \(K_{\text{hist}} = 5\) time steps of server states and predicts one step ahead (\(\Delta = 1\)). 
    The network contains about 0.5M parameters and incurs an average inference delay of 2.1~ms per time slot, which is negligible compared to the overall communication latency.
    
    For ablation analysis, a variant DC-MADDPG is implemented by removing the TimeGNN predictor from TG-DCMADDPG, enabling direct comparison of temporal state forecasting impact on convergence and system efficiency.

    Training is performed for 2000 episodes, and averaged results are reported every 20 episodes. The detailed simulation parameters are summarized in Table~\ref{tab:env_config}.

	\begin{figure}[t]         
		\centering
		\includegraphics[width=1\linewidth]{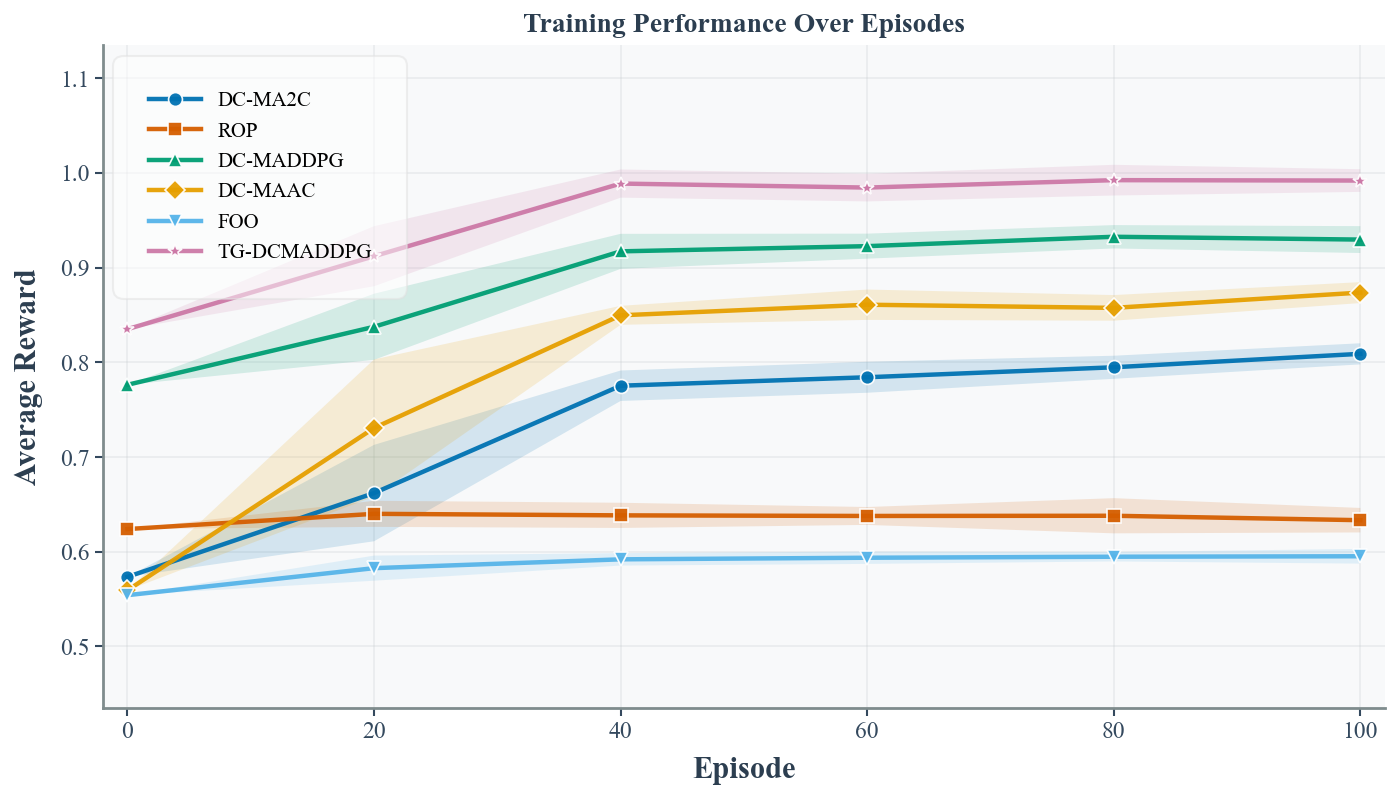}
		\caption{Convergence comparison of average cumulative rewards over episodes.}
		\label{fig:reward_episode}
        \vspace{-2mm}
	\end{figure}

    \begin{figure}[t]         
		\centering
		\includegraphics[width=1\linewidth]{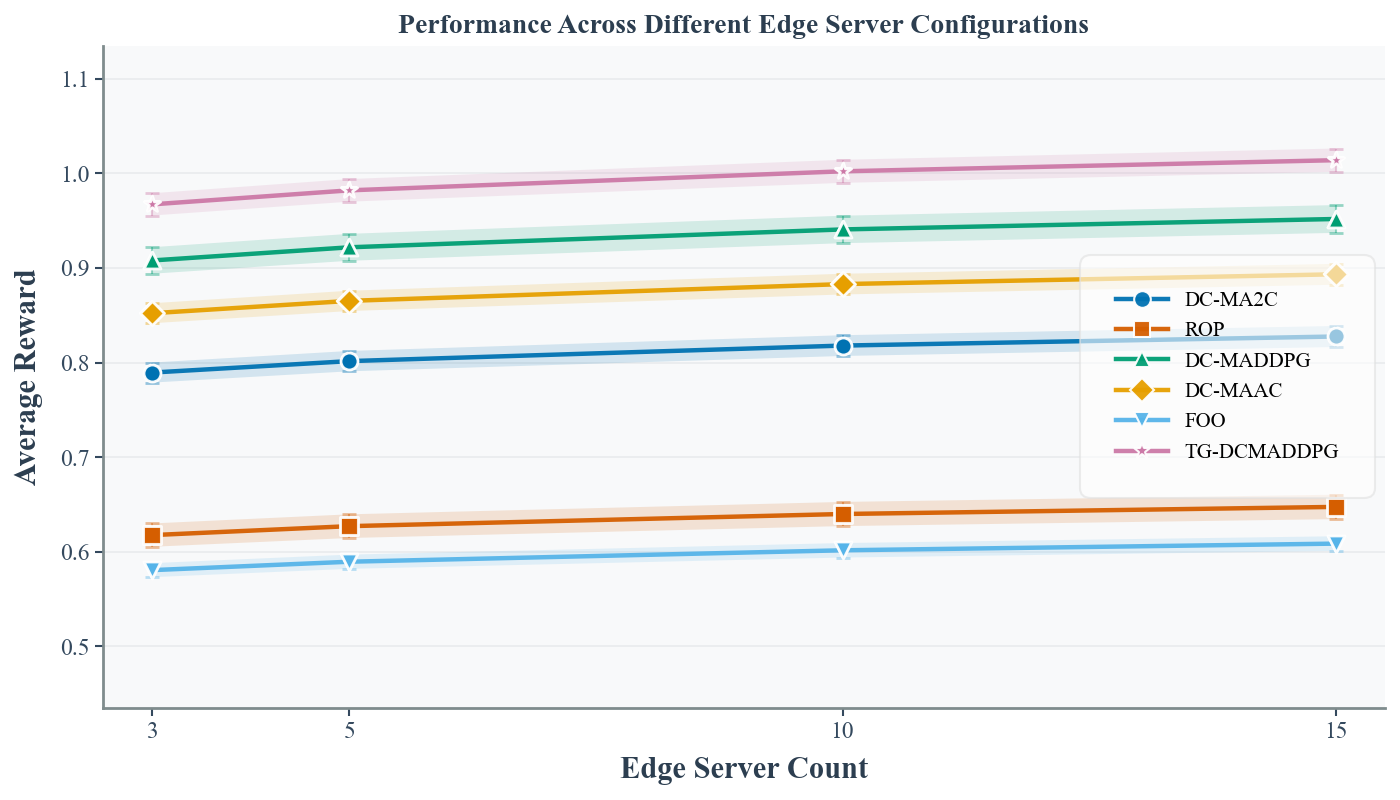}
		\caption{Impact of the number of edge servers on average cumulative rewards.}
		\label{fig:reward_edgeservers}
        \vspace{-2mm}
	\end{figure}

	\begin{table}[htbp]
		\centering
		\caption{Simulation Environment Configuration}
		\label{tab:env_config}
		\renewcommand{\arraystretch}{1.2}
		\begin{tabularx}{0.95\linewidth}{l|X}
			\toprule
			\textbf{Parameter} & \textbf{Value / Range} \\
			\midrule
			Initial task size (MB) & Uniform[30, 50] \\
			Task priority level & Random integer in \{1, 2, 3\} \\
			Edge server CPU load (GHz) & Uniform[4.5, 10] \\
			Bandwidth (MHz) & Uniform[20, 25] \\
			Distance (km) & Uniform[1, 5] \\
			Initial energy of each server (J) & 10{,}000 \\
			Maximum transmission power (dBm) & [0, 50] \\
			Time slots per episode & 10 \\
			Total training episodes & 2000 \\
			Task priority weights & $w_E = 1$, $w_D = 2$ \\
			Actor update frequency & Every 100 steps \\
			Actor learning rate & $1 \times 10^{-3}$ \\
			Critic learning rate & $1 \times 10^{-2}$ \\
			Target network update factor ($\tau$) & 0.01 \\
			Tunable scaling factor ($\beta$) & 0.1 \\
			\bottomrule
		\end{tabularx}
	\end{table}

    \subsection{Convergence Performance Comparison}
    As shown in Fig.~\ref{fig:reward_episode} and Fig.~\ref{fig:reward_edgeservers}, 
    the convergence and scalability of all algorithms are evaluated in terms of average reward. 
    TG-DCMADDPG consistently achieves the highest rewards and the fastest, most stable convergence across all configurations. 
    It quickly exceeds an average reward of 1.0 after about 60 episodes and maintains robust learning efficiency as the number of edge servers increases from 3 to 15. 
    This advantage mainly stems from the TimeGNN-based temporal representation learning, which captures spatiotemporal dependencies under partial observability, 
    and the discrete--continuous hybrid action design, which enables fine-grained coordination between offloading and resource allocation.
    
    In contrast, hybrid-action baselines such as DC-MADDPG, DC-MAAC, and DC-MA2C converge more slowly and saturate at lower reward levels (around 0.85--0.9), 
    especially in denser networks. 
    Their limited ability to model temporal correlations and inter-agent dependencies reduces scalability and adaptability in dynamic MEC environments. 
    ROP shows unstable performance around 0.6 due to random actions, while FOO performs worst because of load imbalance and inefficient resource use. 
    Among learning-based methods, DC-MA2C is more stable but conservative, and DC-MAAC benefits from attention yet lacks temporal awareness.
    
    Overall, these results confirm that integrating temporal graph reasoning with hybrid-action multi-agent reinforcement learning 
    significantly improves convergence speed, scalability, and overall system performance in dynamic MEC scenarios.

	\begin{figure}[t]         
		\centering
		\includegraphics[width=1\linewidth]{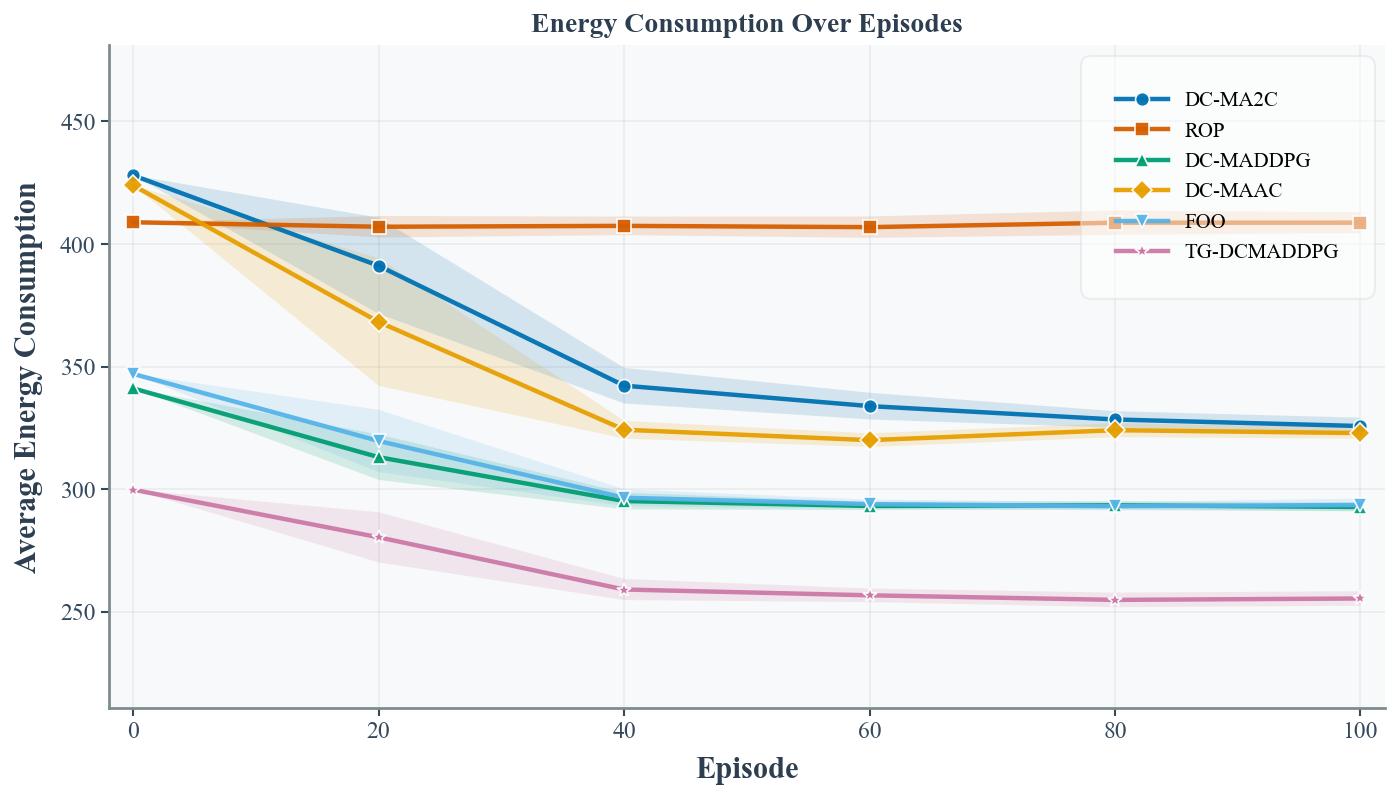}
		\caption{Average energy consumption of different algorithms during training.}
		\label{fig:energy_episode}
	\end{figure}

    \begin{figure}[t]         
		\centering
		\includegraphics[width=1\linewidth]{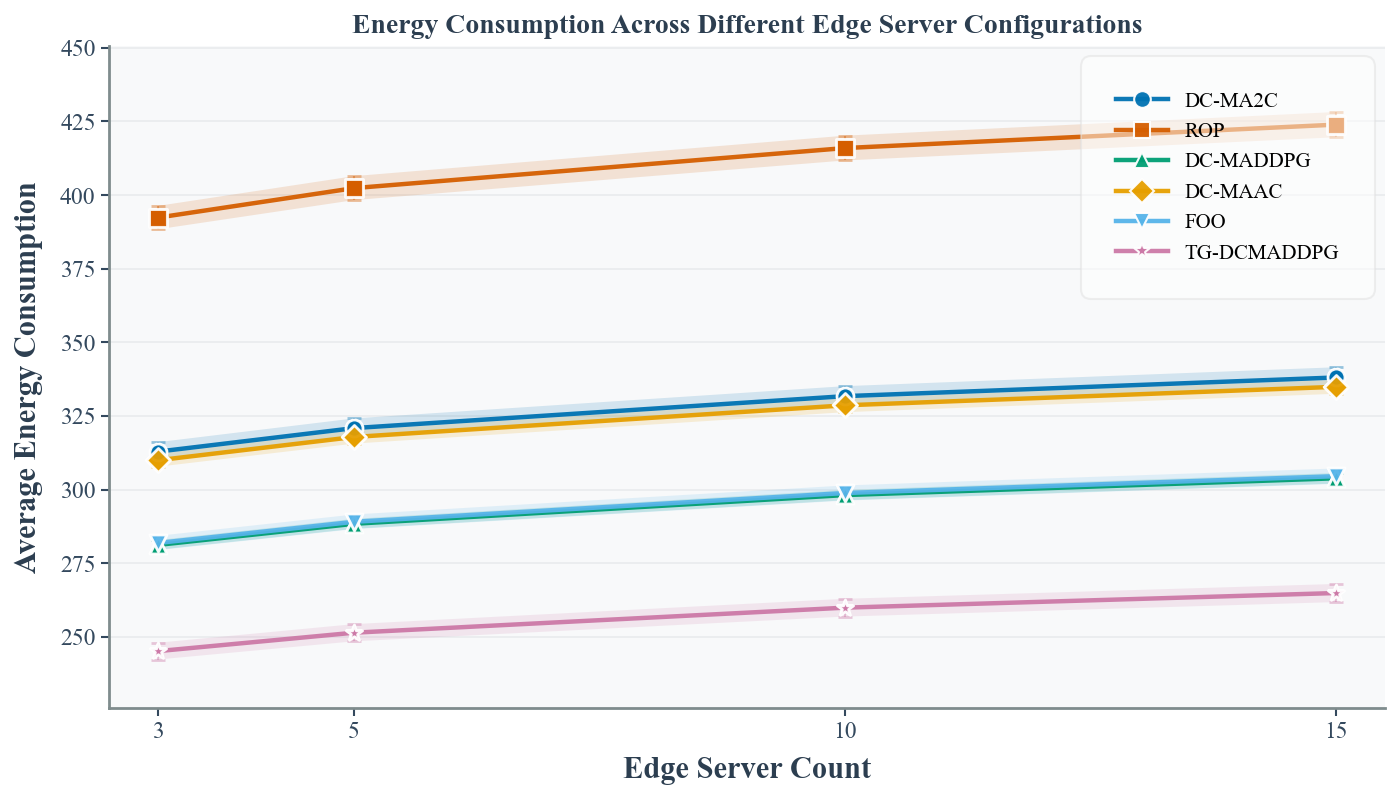}
		\caption{Impact of the number of edge servers on the average energy.}
		\label{fig:energy_edgeservers}
        \vspace{-2mm}
	\end{figure}
	
	\subsection{Energy Efficiency Analysis}
    As shown in Fig.~\ref{fig:energy_episode} and Fig.~\ref{fig:energy_edgeservers}, 
    TG-DCMADDPG and DC-MADDPG consistently achieve the best energy efficiency among all methods. 
    TG-DCMADDPG converges to the lowest energy level, stabilizing below 260 units after about 80 episodes, 
    while DC-MADDPG remains close at around 280 units and maintains stable performance as the network scale increases. 
    As the number of edge servers grows, the overall energy consumption slightly rises due to more frequent communications between devices and multiple edge servers. 
    Nevertheless, TG-DCMADDPG effectively mitigates this impact through its TimeGNN-based predictive modeling, 
    which enhances coordination efficiency and reduces redundant state querying, achieving a balanced trade-off between energy and task performance.
    
    In contrast, DC-MAAC and DC-MA2C consume moderately more energy (310--340 units), 
    and heuristic baselines such as ROP and FOO exceed 400 units due to unoptimized or random offloading decisions. 
    Overall, TG-DCMADDPG maintains high energy efficiency and scalability, 
    effectively balancing predictive accuracy, coordination overhead, and energy-aware decision making in dynamic MEC environments.

	\begin{figure}[t]         
		\centering
		\includegraphics[width=1\linewidth]{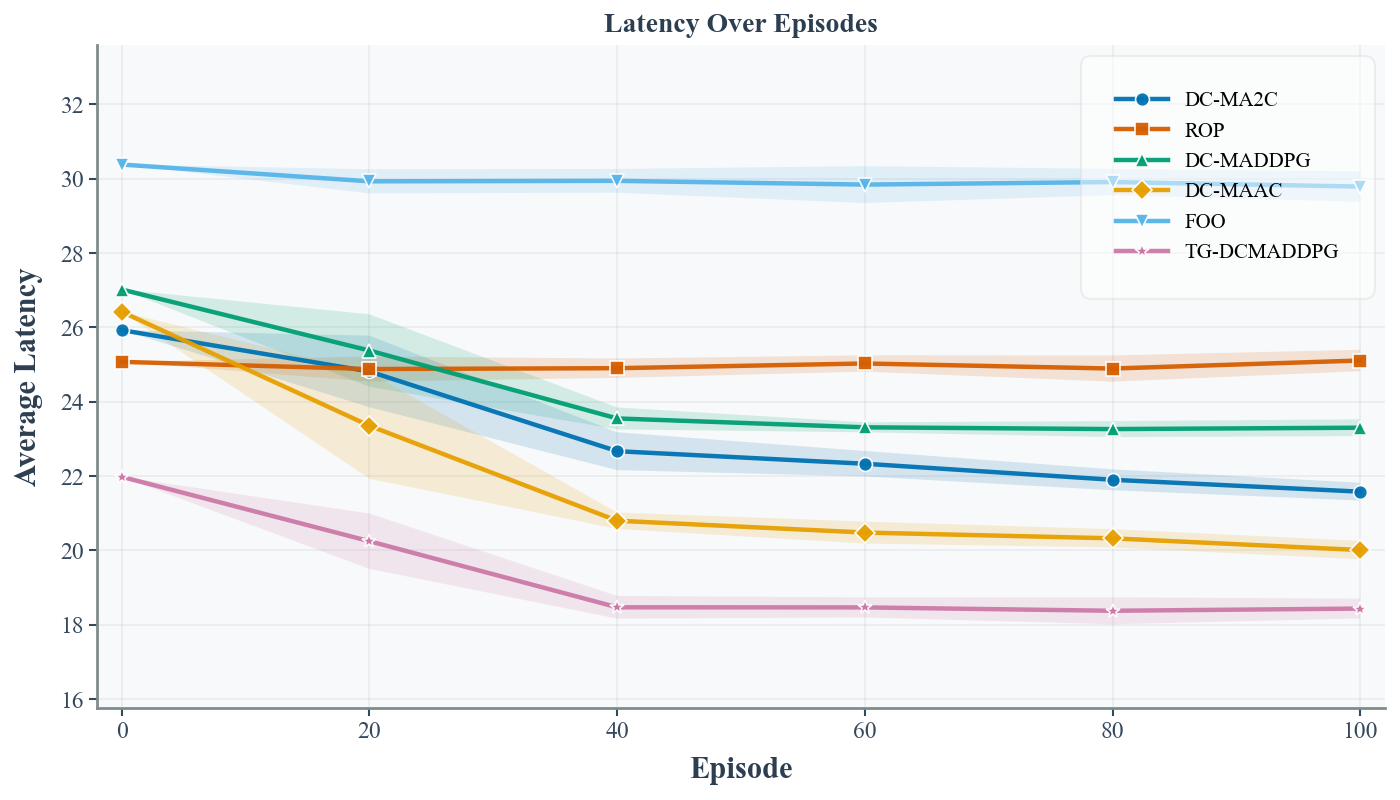}
		\caption{Average task execution latency of different algorithms during training.}
		\label{fig:latency_episode}
	\end{figure}
	
	\begin{figure}[t]         
		\centering
		\includegraphics[width=1\linewidth]{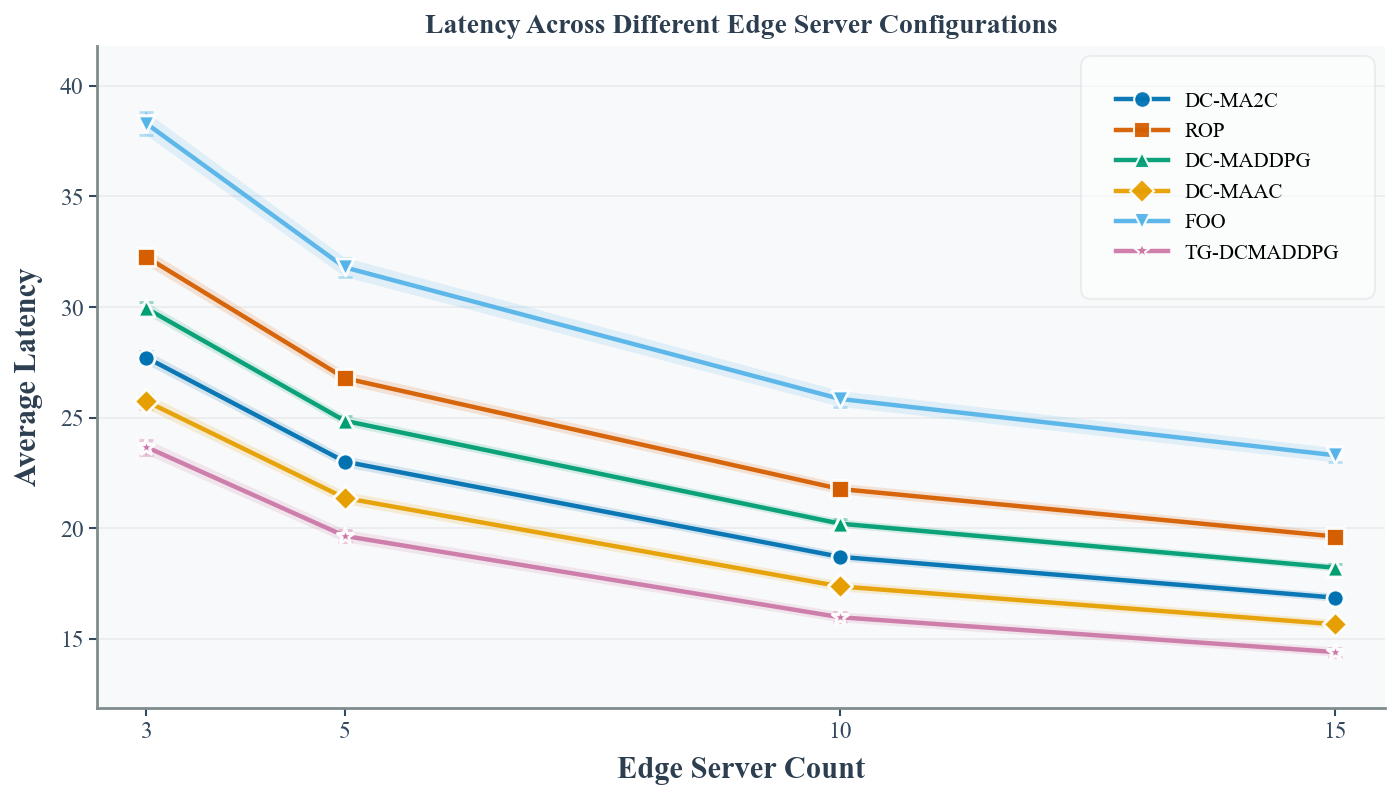}
		\caption{Impact of the number of edge servers on the average task execution latency.}
		\label{fig:latency_edgeservers}
	\end{figure}
    
	\subsection{Latency Performance Analysis}
    As shown in Fig.~\ref{fig:latency_episode} and Fig.~\ref{fig:latency_edgeservers}, 
    TG-DCMADDPG consistently achieves the lowest latency across all network scales, 
    converging rapidly to around 24 units after about 60 episodes. 
    This advantage arises from its TimeGNN-based temporal modeling, 
    which enables proactive prediction of network load and server dynamics for timely and informed offloading decisions.

    In comparison, DC-MAADDPG and DC-MAAC converge to higher latency levels (approximately 26--30 units) 
    due to limited temporal prediction capability, while heuristic methods such as ROP and FOO show much higher latency, 
    often exceeding 32 and 38 units, respectively, because of random or static scheduling. 
    As the number of edge servers increases, the overall latency gradually decreases 
    since computational loads are more evenly distributed and task queuing delays are reduced. 
    TG-DCMADDPG maintains the most stable and lowest latency throughout this process, 
    demonstrating strong scalability and adaptability in dynamic MEC environments. 
    
    Combined with the results in Fig.~\ref{fig:energy_episode} and Fig.~\ref{fig:latency_episode}, 
    TG-DCMADDPG achieves the most balanced performance, maintaining low latency and high energy efficiency. 
    These results highlight the effectiveness of temporal modeling and hybrid-action control in enabling adaptive, 
    scalable, and low-latency decision making for dynamic MEC environments.

	\begin{figure}[t]         
		\centering
		\includegraphics[width=1\linewidth]{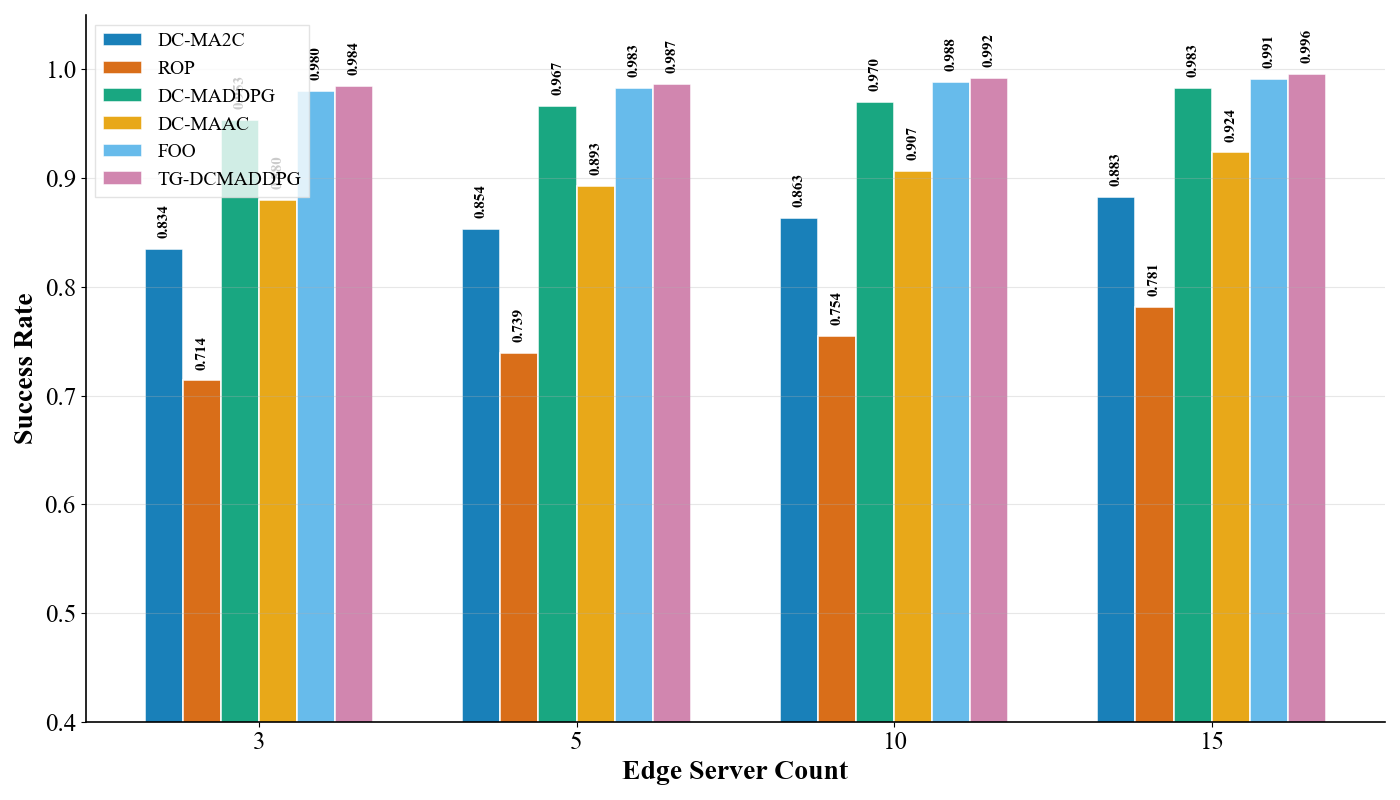}
		\caption{Task completion ratio comparison among all methods.}
		\label{fig:sucess-ratio}
        \vspace{-2mm}
	\end{figure}
	
    \subsection{Task Completion Rate Comparison}

    As shown in Fig.~\ref{fig:sucess-ratio}, 
    the task completion rate of all algorithms is evaluated under different numbers of edge servers. 
    TG-DCMADDPG consistently achieves the highest success rates, exceeding 98\% in all cases and reaching up to 99.6\% with 15 edge servers. 
    This stable performance demonstrates strong scalability and robustness under dynamic conditions, 
    as the TimeGNN-based predictive modeling helps anticipate load variations and balance offloading decisions efficiently. 
    
    FOO attains a similar rate (around 99\%) due to full offloading but incurs higher latency and energy consumption. 
    Among learning-based methods, DC-MADDPG and DC-MAAC achieve moderate results (around 97\% and 92\%), 
    while DC-MA2C performs slightly worse due to limited global coordination. 
    ROP records the lowest success rate (below 80\%) owing to random and non-adaptive actions. 
    
    Overall, TG-DCMADDPG achieves near-perfect task completion and excellent scalability, 
    effectively maintaining reliability and energy–latency efficiency in dynamic MEC environments.

	\section{Conclusion}
	
	In this work, we proposed TG-DCMADDPG, a novel task offloading framework for multi-user mobile edge computing (MEC) environments characterized by energy constraints and dynamic server availability. The proposed framework integrates temporal prediction via a Time-aware Graph Neural Network (TimeGNN) with a discrete-continuous hybrid action multi-agent reinforcement learning algorithm (DC-MADDPG), aiming to balance energy efficiency, task latency, and service reliability.To reduce the frequent energy-consuming state broadcasts from edge servers, especially those lacking stable power sources, we employ TimeGNN to predict future server states using historical status collected every two intervals. This allows the agents to make proactive and informed offloading decisions while significantly extending server operational time.
	


\end{document}